\documentclass[pdflatex,sn-mathphys-num]{sn-jnl}% Math and Physical Sciences Numbered Reference Style 
\usepackage{graphicx}%
\usepackage{multirow}%
\usepackage{amsmath,amssymb,amsfonts}%
\usepackage{amsthm}%
\usepackage{mathrsfs}%
\usepackage[title]{appendix}%
\usepackage{xcolor}%
\usepackage{textcomp}%
\usepackage{manyfoot}%
\usepackage{booktabs}%
\usepackage{algorithm}%
\usepackage{algorithmicx}%
\usepackage{algpseudocode}%
\usepackage{listings}%
\usepackage{booktabs}
\usepackage{tabularx} 
\usepackage{caption} 
\begin{document}

\title[Article Title]{NijiGAN: Transform What You See into Anime with Contrastive Semi-Supervised Learning and Neural Ordinary Differential Equations}


\author*[1,2]{\fnm{Kevin Putra} \sur{Santoso}}\email{kevin@avalon-ai.org}

\author[1]{\fnm{Anny} \sur{Yuniarti}}\email{anny@if.its.ac.id}
\equalcont{These authors contributed equally to this work.}

\author[1,2]{\fnm{Dwiyasa} \sur{Nakula}}\email{5027221001@student.its.ac.id}
\equalcont{These authors contributed equally to this work.}

\author[1]{\fnm{Dimas Prihady} \sur{Setyawan}}\email{5025211184@student.its.ac.id}
\equalcont{These authors contributed equally to this work.}

\author[1]{\fnm{Adam Haidar} \sur{Azizi}}\email{5025211114@student.its.ac.id}
\equalcont{These authors contributed equally to this work.}

\author[1]{\fnm{Jeany Aurellia P.} \sur{Dewati}}\email{5027221008@student.its.ac.id}
\equalcont{These authors contributed equally to this work.}

\author[1]{\fnm{Farah Dhia} \sur{Fadhila}}\email{5025211030@student.its.ac.id}
\equalcont{These authors contributed equally to this work.}

\author[1]{\fnm{Maria T. Elvara} \sur{Bumbungan}}\email{5027211042@student.its.ac.id}
\equalcont{These authors contributed equally to this work.}

\affil*[1]{\orgname{Institut Teknologi Sepuluh Nopember}}

\affil[2]{\orgname{Avalon AI}}


\abstract{Generative AI has transformed the animation industry. Several models have been developed for image-to-image translation, particularly focusing on converting real-world images into anime through unpaired translation. Scenimefy, a notable approach utilizing contrastive learning, achieves high fidelity anime scene translation by addressing limited paired data through semi-supervised training. However, it faces limitations due to its reliance on paired data from a fine-tuned StyleGAN in the anime domain, often producing low-quality datasets. Additionally, Scenimefy's high parameter architecture presents opportunities for computational optimization. This research introduces NijiGAN, a novel model incorporating Neural Ordinary Differential Equations (NeuralODEs), which offer unique advantages in continuous transformation modeling compared to traditional residual networks. NijiGAN successfully transforms real-world scenes into high fidelity anime visuals using half of Scenimefy's parameters. It employs pseudo-paired data generated through Scenimefy for supervised training, eliminating dependence on low-quality paired data and improving the training process. Our comprehensive evaluation includes ablation studies, qualitative, and quantitative analysis comparing NijiGAN to similar models. The testing results demonstrate that NijiGAN produces higher-quality images compared to AnimeGAN, as evidenced by a Mean Opinion Score (MOS) of 2.192, it surpasses AnimeGAN's MOS of 2.160. Furthermore, our model achieved a Frechet Inception Distance (FID) score of 58.71, outperforming Scenimefy's FID score of 60.32. These results demonstrate that NijiGAN achieves competitive performance against existing state-of-the-arts, especially Scenimefy as the baseline model.}




\maketitle

\section{Introduction}\label{introduction}
\begin{figure}[H]
  \centering
  \includegraphics[width=1\textwidth]{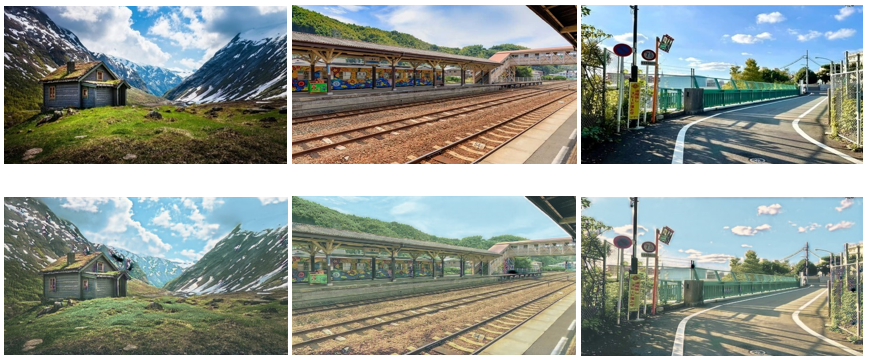}
  \caption{\textbf{Generated samples from NijiGAN}. We show 1366×768 input images (top) and 1366×768 output images that is translated into anime domain (bottom).}
\end{figure}

The significant advancements in Generative Artificial Intelligence (GenAI) \cite{goodfellow2020generative}\cite{karras2019style}\cite{karras2020analyzing} have driven numerous breakthroughs across creative industries, especially in the domains of art and animation. Among the various generative models, Generative Adversarial Networks (GANs) have emerged as one of the most impactful approaches \cite{goodfellow2020generative}\cite{cohen2023generative}\cite{gui2021review}. One of GANs' strengths compared to other models such as diffusion is that GANs enable systems to generate new data that closely approximates the training data distribution with relatively lower computational cost \cite{sharma2024generative}\cite{bengesi2024advancements}. In reality, GAN-based models have seen applications across many domains, such as image transformation, music generation, and text generation \cite{reed2016generative}\cite{zhang2022styleswin}\cite{kim2020remark}. However, challenges remain in applying GANs to highly specific tasks, such as style transfer between visually disparate domains. A notable example of this is unpaired image-to-image translation \cite{zhu2017unpaired}\cite{liu2017unsupervised}\cite{huang2018multimodal}, where the model must learn style transfer between domains without explicitly paired data. Translating real-world landscape images into anime-style images exemplifies this challenge, given the substantial structural and textural differences between the two visual styles \cite{jiang2023scenimefy}.

Various approaches have been developed to address these challenges. For example, CycleGAN employs a cycle consistency mechanism to ensure that transformed images can revert to their original domain without significant loss of information \cite{zhu2017unpaired}. Despite its utility, this approach has limitations, particularly due to its reliance on a bijective mapping that may not be suitable for domains as distinct as real-world and anime images. Moreover, finding high-quality datasets specific to the target domain remains a major challenge, which can result in inconsistent visual quality and style in the generated images. To tackle these obstacles, Jiang et al. introduced Scenimefy \cite{jiang2023scenimefy}, a model designed to enhance anime-style transfer in real-world landscape images through a semi-supervised learning approach. Scenimefy incorporates pseudo-paired data synthesis, semantic segmentation for data selection, and semi-supervised training, yielding more accurate anime style transfer. However, Scenimefy still encounters challenges, such as reducing visual artifacts, minimizing the substantial domain gap between real-world and anime aesthetics, and balancing these improvements with model complexity and inference time.

In this context, we propose NijiGAN, an enhanced model inspired by Scenimefy's semi-supervised training framework, with several modifications to improve image quality and reduce the computational complexity on scenimefy so that translating real image to anime in real time become much possible. NijiGAN utilizes Neural Ordinary Differential Equations (NeuralODEs) as a replacement for traditional ResNet layers in Scenimefy’s generator, which aims to decrease model complexity while preserving or enhancing visual quality. In summary, our contributions to NijiGAN include:
\begin{itemize}
\item We introduce NijiGAN, a new framework for anime style transfer that uses Neural Ordinary Differential Equations (NeuralODEs) within the CUT generator from Scenimefy. NeuralODEs layers make NijiGAN smaller and faster without sacrificing image quality, making it better suited for real-time rendering.
\item We compare NijiGAN with the original CUT generator to show how replacing ResNet blocks with NeuralODEs layers impacts efficiency and image quality.
\item To support further research in scene stylization, we also add a high-resolution anime scene dataset which aims to enhance future developments in anime-style generative modeling. 
\end{itemize} 

\section{Related Works}\label{related-works}

\subsection{Scenimefy}\label{scenimefy} 
Scenimefy tackles the challenge of turning real images into anime by handling the unique complexities of anime style, limited high-quality datasets, and the big gap between real and animated visuals. To make this work, Scenimefy blends supervised and unsupervised learning, which helps the model stay consistent by using pseudo-paired data and unpaired data, containing the anime style references. The process has three main steps: creating pseudo-paired data, filtering pseudo-paired data based on semantic segmentation, and semi-supervised training. The pseudo-paired data helps in supervised training, so the model learns to keep essential details during the transition. Semantic segmentation filters out low-quality images that do not match the needed structure. This approach allows the model to pick up both the structure of real-world images and the distinct anime style. On the early stages, the model relies more on guided training, but as it progresses, it gradually learns to handle the task with less supervision \cite{jiang2023scenimefy}.

\subsection{Neural Ordinary Differential Equations}\label{neuralodes} 
NeuralODEs treat each layer of a neural network as a continuous transformation, rather than a series of discrete layers such in ResNets. This approach can reduce model complexity and, in some cases, produce smoother and more consistent outputs. By learning continuous dynamics, NeuralODEs help models generalize better, especially in tasks like image transformation where smoothness and fine detail are crucial \cite{Khrulkov_2021_ICCV}. In NijiGAN, NeuralODEs replace the ResNet bottleneck on Scenimefy that employs traditional CUT model. It aims to simplify the model without sacrificing quality or detail, especially when translating complex scenes from real-world landscapes to anime.

\subsection{Contrastive Learning for Unpaired Image-to-image Translation}\label{contrastivelearning}
Contrastive learning has become a popular method for achieving style transfer without requiring paired datasets \cite{LEE2024111170}\cite{10.1007/978-3-030-58545-7_19}\cite{Han_2021_CVPR}. Methods like CUT (Contrastive Unpaired Translation) use contrastive learning to match patches of translated images to the original domain while maximizing the difference with other images, which maintains coherence across unpaired domains. CUT has shown impressive results in image-to-image translation tasks, preserving key features while adapting style \cite{jiang2023scenimefy}. NijiGAN builds on this by integrating contrastive loss functions that enhance anime style consistency, ensuring translated images maintain the intricate look and feel of the anime style references. By refining these techniques within a semi-supervised framework, NijiGAN aims to bridge the real-to-anime domain gap while minimizing artifacts.

\section{Methodology}\label{methodology}

\begin{figure}[htbp]
  \centering
  \includegraphics[width=1\textwidth]{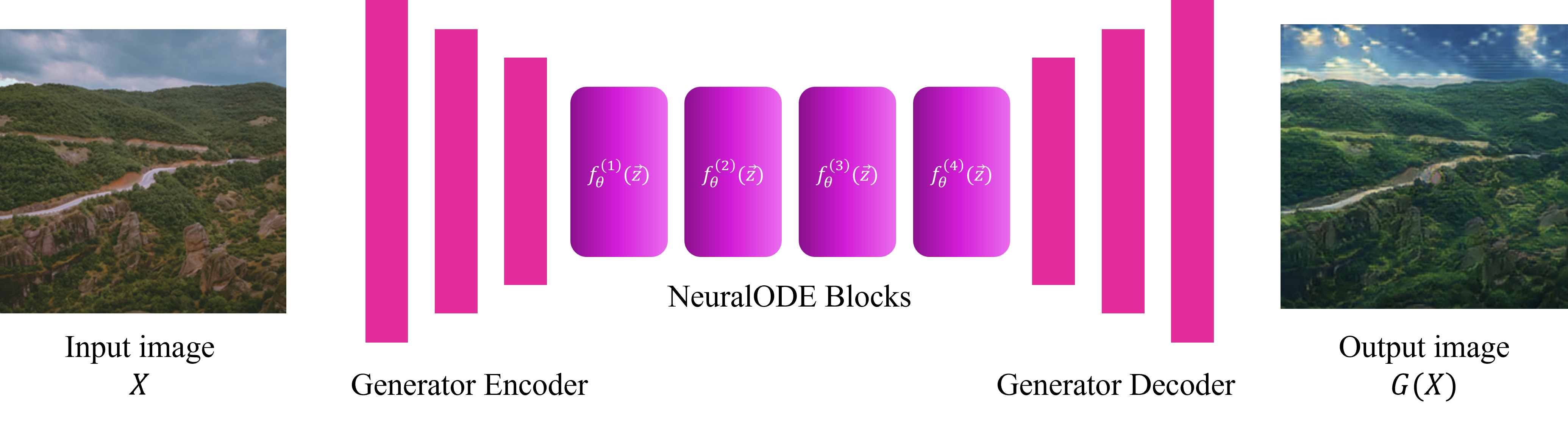}
  \includegraphics[width=1\textwidth]{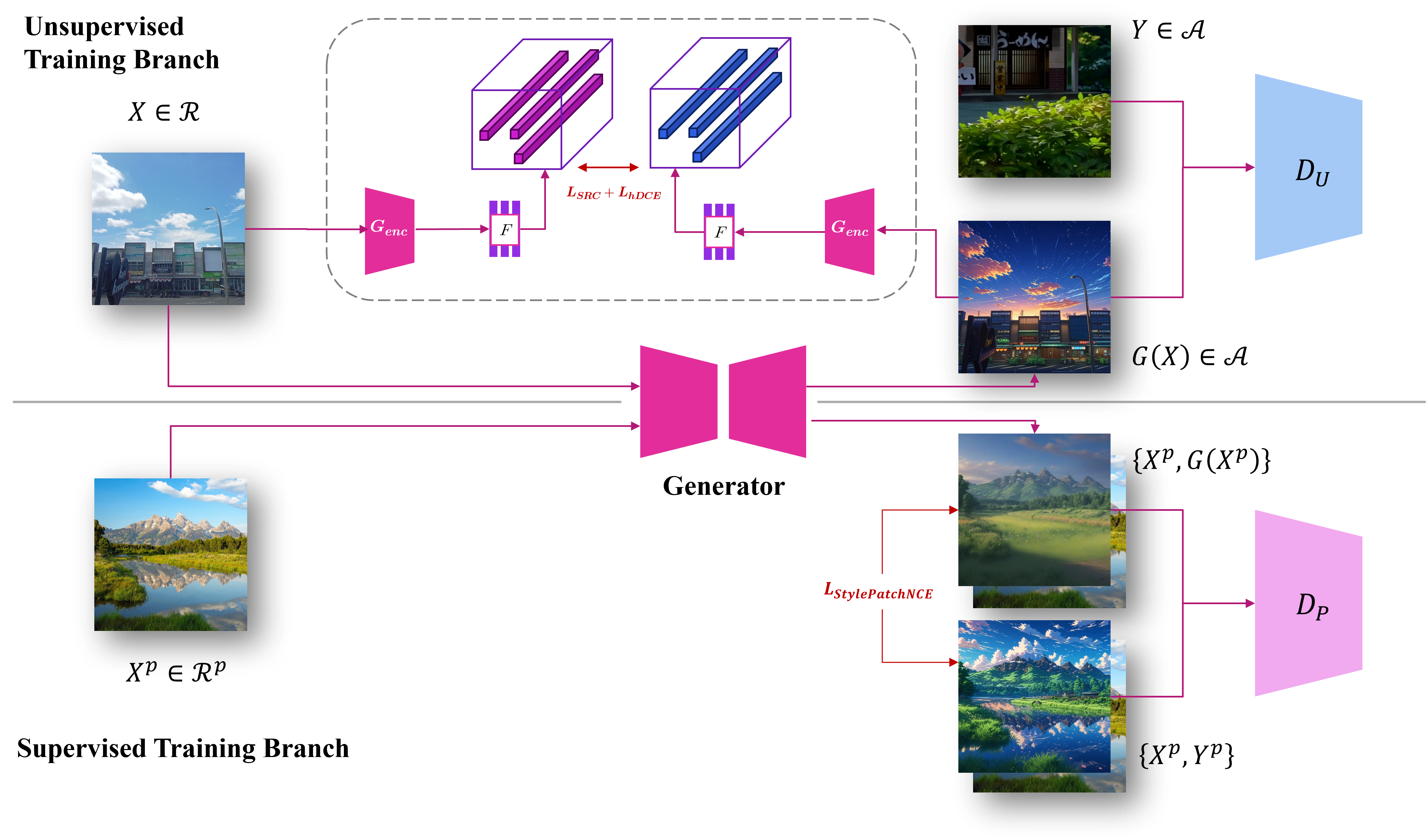}
  \caption{\textbf{The Architecture of NijiGAN}. We propose the use of NeuralODEs in the generator as a replacement for the CUT generator, which implements ResNet blocks. Through the use of NeuralODEs, we employ a continuous dynamics method during the training process.}
\end{figure}

Our goal is to transform real-world images from the input domain to appear like anime images from the output domain. To achieve this, we prepare several types of paired datasets: pseudo-paired data $ \{(x_p, y_p)\mid{x_p\in{X^p}, y_p\in {Y^p}}\}$ and unpaired data $\{(x,y)\mid{x\in{X}, y\in{Y}}\}$. The pseudo-paired data is generated by translating collections of real-world images to anime using the pretrained Scenimefy model to ensure better structural and anime style consistency, especially with respect to the anime style reference. The purpose of the pseudo-paired data is to provide information about the target of the translated images so that the generator can follow the characteristics of its pseudo-target, namely the anime style. In this case, $Y^p$ has textures and semantic features resembling anime, and these features will be used for the generator's learning process. In other words, pseudo-paired data facilitates the supervised learning process. Meanwhile, unpaired data aims to provide information about the reference painting style. In this case, Makoto Shinkai's paintings are adopted as the main reference. For more details regarding the preparation of paired and unpaired data, please refer to Jiang et al. \cite{jiang2023scenimefy}. 

\subsection{Neural Ordinary Differential Equations in CUT}\label{neuralode-in-cut}
In general, CUT consists of three components: encoder, bottleneck, and decoder. The encoder functions to extract important features from the input images in domain $X$ or $X^p$. In the CUT architecture, the encoder uses a series of convolutional layers to transform the input images into more abstract latent representations. This representation captures essential semantic and texture information required for the style translation process. The bottleneck then takes the latent representation from the encoder. In CUT, the bottleneck plays a crucial role in ensuring that the features extracted from the input domain $X$ or $X^p$ can be effectively mapped to the output domain $Y$ or $Y^p$ —an effective discriminative representation for the translation task without requiring directly paired data. The decoder is responsible for reconstructing images from the latent representation processed by the bottleneck. The decoder transforms these features into output images in domain Y or $Y^p$, in this case, the anime style. The decoding process must ensure that the original content of the input image is preserved while its visual style is changed according to the target. In NijiGAN, we implement NeuralODEs to replace the discrete ResNet layers in the bottleneck so that the evolution of the state in each layer of the neural network can proceed dynamically.

NeuralODEs is a differential equation that utilizes neural networks to parameterize vector spaces. Generally, NeuralODEs is defined by the following formula:

\begin{equation}
    \frac{d\textbf{h}}{dt}(t) = f_{\theta}(t, \textbf{h}(t)); \textbf{h}(0) = \textbf{h}_0
\end{equation}

where $f_{\theta}(t, \textbf{h}(t))$ is a layer composed of Convolutional Neural Networks, Activation Functions, and Batch Normalizers. This formula is obtained through modeling a ResNet:

\begin{equation}
    \textbf{h}_{t+1}=\textbf{h}_t + f(\textbf{h}_t,\theta_t ),
\end{equation}

where $\textbf{h}_t$ represents the hidden state of the neural network at layer $t$. Note that the state is updated discretely by adding the output of a function that depends on the hidden state and parameter $\theta_t$. If each state change is defined with a step size variable $\Delta t$, then the equation above will resemble the Euler method, a numerical technique for solving differential equations:

\begin{equation}
    \textbf{h}_{t+1} = \textbf{h}_t + \Delta\textbf{h} = \textbf{h}_t + \Delta t \cdot f(\textbf{h}_t,\theta_t)
\end{equation}

Here, $f(\textbf{h}_t,\theta_t)$ is a function that defines how the hidden state changes using parameter $\theta_t$. Since $\Delta h = \textbf{h}_{t+1} - \textbf{h}_t = \Delta t \cdot f(\textbf{h}_t, \theta_t)$, we can write it as

\begin{equation}
    \frac{\textbf{h}_{t+1} - \textbf{h}_t}{\Delta t} = f(h(t), \theta_t)
\end{equation}

The Euler method approximates the solution of an ODE by discretizing the time into small steps $\Delta t$. If $\Delta t$ becomes very small, the update of $\textbf{h}_{t+1}$, which initially appears discrete because it has discrete time segments $[t_0,t_1,t_2,…]$ with $t_{k+1}=t_k+\Delta t$, will appear continuous, forming the interval $[t_0,t_k ]$. In other words, its limit approaches zero, so the hidden state will have dynamics of a function dependent on time:

\begin{equation}
    \lim_{\Delta t \to 0} \frac{\textbf{h}(t + \Delta t) - \textbf{h}(t)}{\Delta t} = \frac{d\textbf{h}}{dt}(t)
\end{equation}

Thus, the differential equation in equation 1 is formed, and the modeling of layers in the neural network is now defined continuously. If there is a range that describes the depth of this neural network, say $[t_0,t_1 ]$, then the solution to this differential equation is

\begin{equation}
    \textbf{h}(t_1) = \textbf{h}(t_0)+\int_{t_0}^{t_1} f_\theta (t,\textbf{h}(t))dt
\end{equation}

To solve this differential equation, numerical techniques are needed, for example, the Runge-Kutta method. However, NijiGAN applies the Dormand-Prince method (RKDP), a family of the Runge-Kutta methods \cite{yan2019robustness}\cite{chen2018neural}\cite{kimura2009dormand}. Semi-supervised learning is employed to map an image $x$ from the real-world domain $R$ to an image $y$ in the anime domain $A$ by leveraging a pseudo-paired dataset $P = \left\{ x_{p}^{(i)},y_{p}^{(i)} \right\}_{i=1}^{N}$. In general, this training process utilizes two branches: supervised learning branch and unsupervised learning branch.

\textbf{The supervised branch} is implemented to enable the model to learn the extracted features from a synthetically generated pseudo-paired dataset. These extracted features will be utilized in the training process and in mapping features for scene stylization. This supervised branch is based on the conditional GAN framework \cite{jiang2023scenimefy}, which is trained using a conditional adversarial loss as follows:

\begin{equation}
    \mathcal{L}_{cGAN}(G, D_P) = \mathbb{E}_{{y^p}, {x^p}}\left[ \log{D_P({y^p}, {x^p})}\right] + \mathbb{E}_{x^p}\left[ \log{(1-D_P({x^p}, G(x^p))} \right]
\end{equation}

Where the patch discriminator $D_P$ is used to distinguish between $\{(y^p,x^p )\}$ and $\{(G(x^p ),x^p)\}$, and $(\cdot,\cdot)$ denotes the concatenation operation.

In this supervised training process, instead of using ground truth from the target domain $Y$, we utilize pseudo-ground truth $Y^p$. We apply patch-wise contrastive learning to help the model learn robust local similarity and focus on fine details. This approach maps patches from the generated image to their corresponding patches in the pseudo-ground truth, providing robust supervision. This mapping leverages similar patches between the translated image and its pseudo-ground truth, rather than relying on identical patches. Patches located in the same position should be embedded closer together, while those in different locations should be kept further apart. By using the StylePatchNCE loss \cite{jiang2023scenimefy}, we split the generator into two components, the encoder $G_{enc}$  and the decoder $G_{dec}$. The features from  $G_{enc}$  facilitate image translation by aligning patches that match the input image. These features are subsequently fed into a two-layer trainable MLP network $F$ to obtain the embedded patch features. To capture anime texture at different levels of granularity, we extract multi-scale features from a total of $L$ layers in $G_{enc}$. The StylePatchNCE is defined as follows \cite{jiang2023scenimefy}:

\begin{equation}
    L_{StylePatchNCE}(G, F, Y^p) = \sum_{l=1}^{L} \sum_{i \neq j}L_{patch}^{style}(\tilde{v}_l^i , v_l^i , v_l^j)
\end{equation}

Where $\tilde{v}_l^i$ and $v_l^i$ are the $l$-th-layer embedded patch at the location $i$ of $G(x^p)$ and $y^p$, respectively, and $L_{patch}^{style}$ is as follows:

\begin{equation}
    L_{patch}^{style} = -\log{\left[ \frac{\exp{(v \cdot v^+)}}{\exp{(v \cdot v^+)} + \sum_{i=1}^N \exp{(v \cdot v_i^-)}}  \right]}
\end{equation}

$L_{patch}^{style}$ allows the model to preserve the structure of the original image during translation to a different domain by maximizing the similarity between the relevant data pairs $\exp{(v \cdot v^+)}$ and minimizing the similarity between irrelevant pairs $\exp{(v \cdot v_i^-)}$ \cite{jiang2023scenimefy}.

Overall, the total loss for the supervised training branch is formulated as:

\begin{equation}
    L_{sup} = L_{cGAN}(G, D_P) + \lambda_{style}L_{StylePatchNCE}(G, F, Y^p)
\end{equation}

where $\lambda_{style}$ is weight for StylePatchNCE loss.

\textbf{The unsupervised branch} is designed to replicate the visual style of the Makoto Shinkai anime dataset in domain $A$. In this branch, the goal is to adapt anime images to the Makoto Shinkai style while preserving the original structure. In the real world, image patches from various locations have highly heterogeneous semantic relationships, making it challenging to find matching patches for training. However, these diverse semantic relationships must be considered to achieve image translation that aligns with the content \cite{jung2022exploring}. To address this, a semantic relation consistency loss $L_{SRC}$ is applied to minimize the Jensen-Shannon Divergence (JSD) between the similarity patch distributions in domain $x$ and $G(x)$ \cite{jung2022exploring}.

\begin{equation}
    L_{SRC} = JSD \left( G_{enc}(x) \mid{G_{enc}(G(x))} \right).
\end{equation}

Additionally, a hard negative contrastive loss $L_{hDCE}$ is necessary to progressively enhance the ability to identify hard negatives by designing the distribution of hard negatives and establishing connections with the similarity relations \cite{jung2022exploring}:

\begin{equation}
    L_{hDCE} = \mathbb{E}_{z,w\sim{p_{ZW}}} \left[ -\log{\left( \frac{\exp{(w^T z/\tau)}}{(N\mathbb{E}_{z^- \sim{q_{z^-}}} [\exp{(w^T z^-/\tau)}])} \right)} \right],
\end{equation}

where $N$ negatives and $\tau$ is the temperature parameter for each positive pair $(z,w)\sim{p_{ZW}}$ with z being the embedded vector from patch $Z$ in the synthetic anime image and $w$ the embedded vector from patch $W$ in the anime style reference image. Thus, the total loss for the unsupervised training branch is formulated as:

\begin{equation}
    L_{unsup} = L_{GAN}(G, D_U) + \lambda_{SRC}L_{SRC} + \lambda_{hDCE}L_{hDCE},
\end{equation}

where $D_U$ is discriminator for unsupervised training, $\lambda_{SRC}$ is weight for semantic relation consistency loss, and $\lambda_{hDCE}$ is the weight for hard negative contrastive loss.

\subsection{Overall training}\label{overalltraining}
The Adam optimizer is used to optimize the parameters of NijiGAN based on the loss function $L_{i2i}$ during the backpropagation phase of training. The overall training scheme consists of 25 epochs, with each epoch having 24,000 steps. The first 10 epochs serve as warm-up epochs. After the initial 10 epochs, the supervised weight loss, $\lambda_{sup} (t)$, is gradually applied as the epochs progress. 
The total loss for one epoch is formulated as:
\begin{equation}
L_{i2i}=L_{unsup} + \lambda_{sup} L_{sup}
\end{equation}

where the supervised lambda $\lambda_{sup}(t)=\cos{\frac{\pi}{40} (t-1)}$ with $t$ representing the $t$-th epoch.

\section{Results}\label{results}

\subsection{Dataset}\label{dataset}
The training process is divided into two branches. Supervised training is conducted to train the model to learn the mapping of semantic features from real-world images and construct features from the anime domain into a full image. Unsupervised training is conducted to train the model to understand the color and style distribution of the reference anime images. In this research, we adopted scenes from Makoto Shinkai's anime as reference anime images. To perform semi-supervised training, two kinds of dataset pairs are built: the pseudo-paired dataset and unpaired dataset. In the pseudo-paired dataset, we created 24000 anime images that were translated from real-world to anime images using Scenimefy, and then reduced their size to 256x256. As a result, there are 24000 original images from LHQ dataset and 24000 anime images translated from LHQ dataset using Scenimefy. Our research utilizes scenes from several Makoto Shinkai's anime, similar to the approach used in Scenimefy. We used five anime: 5 Centimeters per Second, Children Who Chase Lost Voices, Your Name, Weathering with You, and Suzume. A total of 4,821 images were extracted from these works, with irrelevant images removed. These images serve as the target dataset for our unpaired training, with the source dataset comprising 10,000 randomly sampled images from the LHQ dataset.

\subsection{Evaluation metrics}\label{evaluation-metrics}
To analyze the performance of the model and the quality of the images produced, we employ an integrated quantitative and qualitative approach. The quantitative approach aims to obtain an exact value that reflects how well the model produces images according to expectations. In this case, we use the Frechet Inception Distance (FID) as the primary metric to numerically evaluate the image characteristic based on the anime style reference. Meanwhile, a qualitative approach is used to assess image quality based on human observation. This is done so that the assessment can provide a more subjective and contextual perspective. To bridge these two approaches, we use Mean Opinion Score (MOS) calculation, which is both semi-quantitative and semi-qualitative. The MOS metric is derived from the subjective judgments of several human evaluators, who compare images with identical content generated by the NijiGAN model and other comparison models. This process enables us to obtain a holistic understanding of the quality of the images produced by the model, both in terms of objective numbers and subjective judgments.

\subsubsection{Quantitative Result}\label{quantitative-result}
Table \ref{FIDandMOS} shows the quantitative evaluation of our proposed method compared to baseline approaches. Our method achieves the lowest Fréchet Inception Distance (FID) score, indicating superior quality in the translated images, which aligns with the higher visual quality observed in our qualitative assessments. Additionally, we compute the FID between the real-world and anime scene datasets as a reference point. The style distribution of our generated results is significantly closer to the anime domain than to the real-world images.

In addition to analyzing the FID metric, we carried out a user study involving 30 participants to assess the quality of anime scene rendering based on three aspects: clear anime-style representation, accurate semantic content consistency, and overall rendering performance. Participants were asked to select the best results from six different methods across 10 image sets. The average preference scores, detailed in Table \ref{FIDandMOS}, indicate that NijiGAN performs similarly to Scenimefy, further highlighting the effectiveness of our approach.

\begin{table}[hbtp]
\caption{\textbf{Comparison of FID and MOS scores.} FID serves as a quantitative evaluation metric that measures the similarity of feature distributions between translated images and reference images, specifically anime scenes. A lower FID score indicates higher quality of the translated images in relation to anime scenes. On the other hand, we conducted a semi-quantitative evaluation using MOS, which is based on user preferences. A higher MOS score reflects a greater user preference for the translated images.}\label{FIDandMOS}%
\begin{tabular*}{\textwidth}{@{\extracolsep\fill}lcccccc}
\toprule
Method & CartoonGAN \cite{andersson2020generative}  & AnimeGAN \cite{chen2020animegan} & Scenimefy \cite{jung2022exploring} & Ours\\
\midrule
FID $\downarrow$    & 45.79   & 56.88  & 60.32
 & 58.71  \\
 MOS $\uparrow$    &  2.708   & 2.160
  & 2.232
 & 2.192 \\
\botrule
\end{tabular*}
\end{table}

\begin{table}[hbtp]
\caption{\textbf{The comparison score between NijiGAN and Scenimefy}. The results indicate that NijiGAN outperforms in optimizing memory allocation and model parameters, as shown by its lower peak memory usage and smaller model parameter values compared to Scenimefy. However, NijiGAN remains less efficient in terms of average computation time and TFLOPS, as shown by its higher average computation time and TFLOPS scores. This inefficiency is caused by the use of continuous dynamics methods during training, which lead to increased model complexity. }\label{NijiGAN vs Scenimefy}
\begin{tabularx}{\textwidth}{@{\extracolsep\fill}lXXXXX}
\toprule
Model & Avg. Five Computation Time (s) & Avg. Fifty Computation Time (s) & Peak Memory (GB) & Model Parameters (M) & TFLOPS \\
\midrule
NijiGAN (Ours) & 1.533 & 0.771 & 9.694 & 5.477 & 0.771 \\
Scenimefy \cite{jiang2023scenimefy} & 1.253 & 0.387 & 9.718 & 11.378 & 0.387 \\
\bottomrule
\end{tabularx}
\end{table}

\subsubsection{Qualitative Result}\label{qualitative-result}
\begin{figure}[h]
  \centering
  \includegraphics[width=1\textwidth]{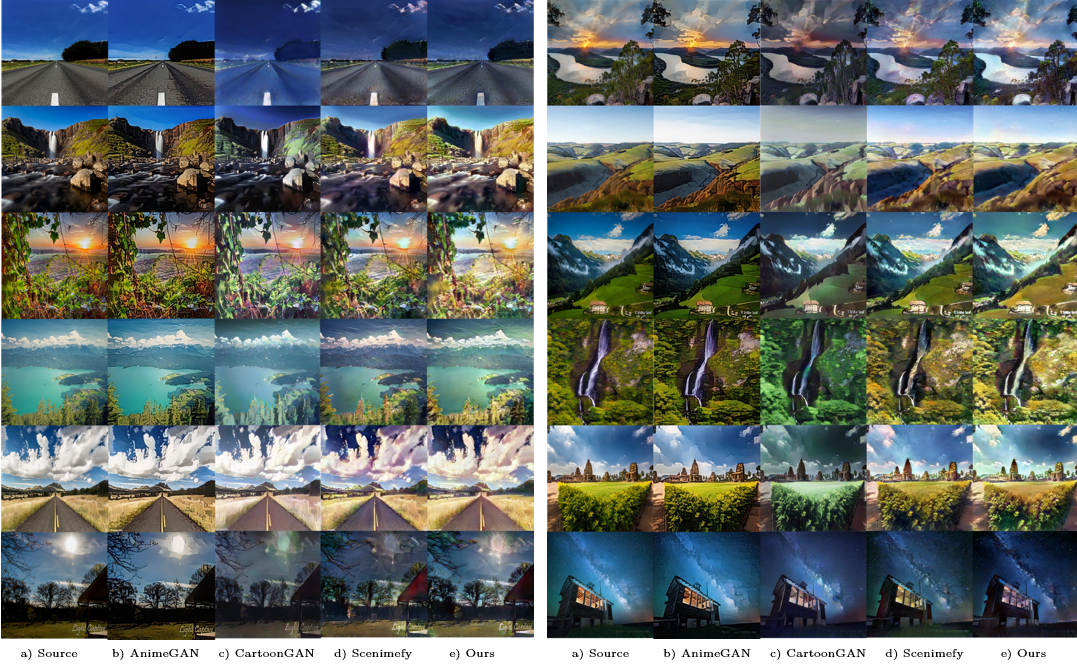}
  \caption{\textbf{The quality images comparison of NijiGAN and several baseline models.} Results of five sample source images stylized by b) AnimeGAN \cite{chen2020animegan}, c) CartoonGAN \cite{andersson2020generative}, d) Scenimefy \cite{jiang2023scenimefy} and e) our method respectively.}\label{generated-image-comparison}
\end{figure}

Figure \ref{generated-image-comparison} presents a qualitative comparison among all baseline models. NijiGAN demonstrates competitive results when compared to these baseline models. Across our provided samples, NijiGAN achieves a balanced representation between the original content and the adopted anime stylistic structures. NijiGAN successfully generates images with refined structures reminiscent of anime paintings. In contrast, CartoonGAN and AnimeGAN employ manually crafted anime-specific losses, such as edge-smoothing loss, in an attempt to achieve sharp edges, which ultimately constrains the extent of style transfer. Furthermore, AnimeGAN produces results that appear more like textural abstractions with weaker stylization, barely exhibiting any anime-style characteristics. While Scenimefy achieves effective style transfer in several samples, it exhibits notable artifacts, including incorrect day-to-night conversion and the introduction of textural elements that lead to image blurring in certain areas. NijiGAN, on the other hand, consistently delivers stable and smooth results.

\subsubsection{Ablation study}\label{ablation_study}

\begin{figure}[h]
  \centering
  \includegraphics[width=1\textwidth]{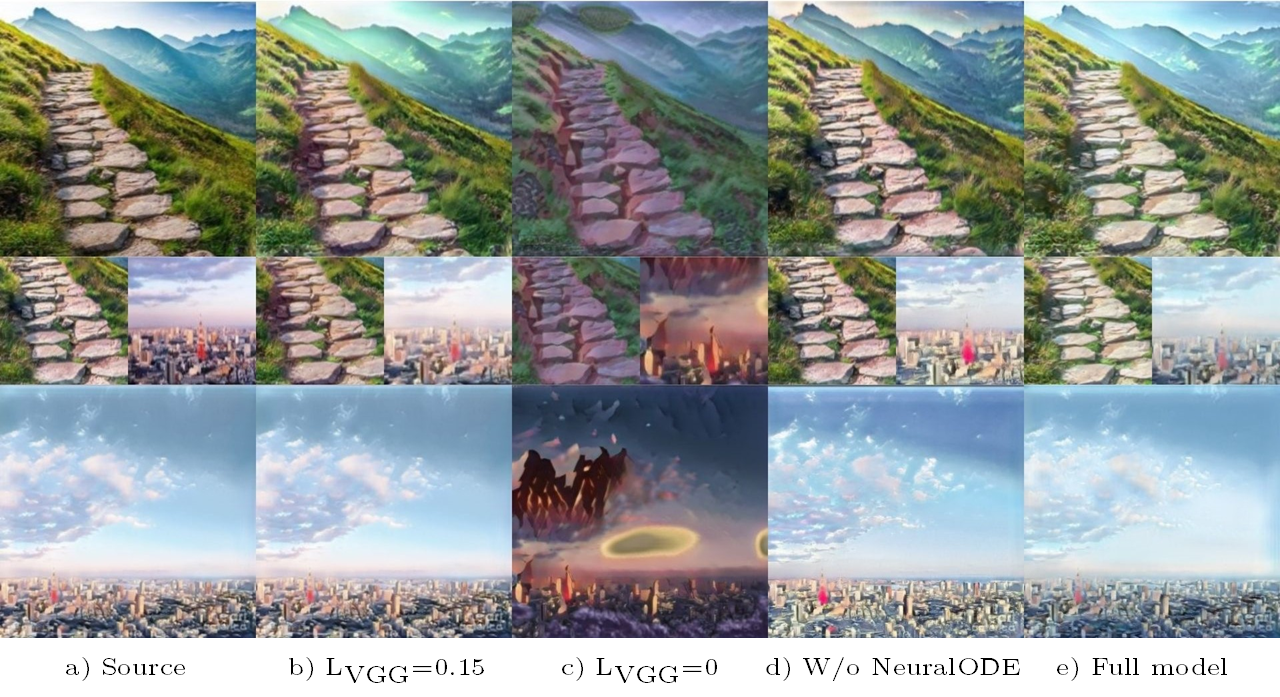}
  \caption{\textbf{Ablation study of several key components in NijiGAN.} We illustrate the different effect of each key components to the translated images.}\label{ablationstudy}
\end{figure}

We also present comparative analyses of component contributions within the NijiGAN model in Figure \ref{ablationstudy}. The scenarios presented include model conditions under: 1.) Implementation with VGG Loss at 0.15, 2.) Absence of VGG Loss, 3.) Replacement of NeuralODE with ResNet, and 4.) The complete model architecture (VGG Loss at 0.1).
The results indicate that VGG loss plays a crucial role in image generation. Models trained with VGG loss demonstrate superior structural consistency, though they tend to eliminate fine details and many anime characteristics from the source images. Additionally, the implementation of NeuralODE has proven effective in generating images with enhanced smoothness characteristics.

\section{Discussion}\label{discussion}
The model we trained, NijiGAN (ours), was evaluated using the Frechet Inception Distance (FID) metric, which measures the similarity between the images generated by the model and the reference images, in this case, anime-style images. As presented in Table \ref{FIDandMOS}, NijiGAN achieved an FID score of 58.71, which is lower than the benchmark model, Scenimefy (4 ResNet blocks). This demonstrates that NijiGAN can produce images with a distribution closer to the reference anime-style images compared to Scenimefy. However, in certain samples, the translated images generated by NijiGAN still lack anime-specific characteristics in terms of texture and strokes. Despite this, NijiGAN exhibits greater consistency and stability than Scenimefy, as evidenced by the background of the translated images, which remains consistent with the original images.

\begin{figure}[H]
  \centering
  \includegraphics[width=1\textwidth]{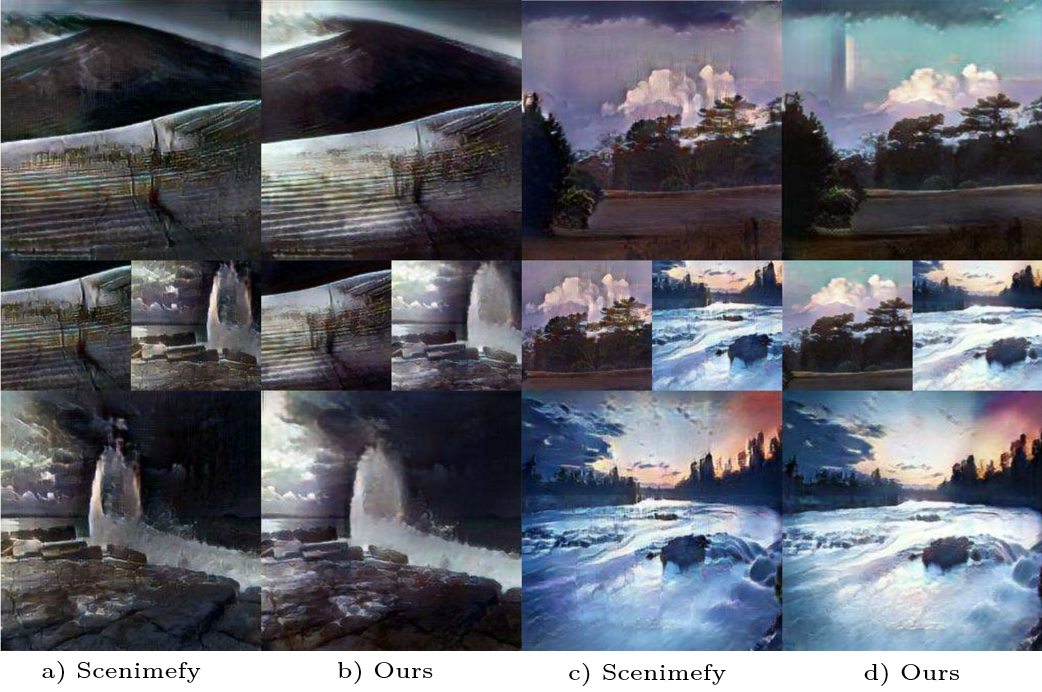}
  \caption{\textbf{NijiGAN (Ours) and Scenimefy (4 ResNet blocks) quality images comparison.} NijiGAN eliminates the checkered artifacts, the translated images are relatively smooth than Scenimefy.}\label{checkeredartifacts}
\end{figure}

Additionally, NijiGAN successfully optimizes memory usage due to its implementation of NeuralODE, which employs continuous dynamics during training. The use of NeuralODE as a replacement for ResNet has proven to be an effective solution for optimizing the number of parameters in the model while maintaining high-quality image translation. This is evident from NijiGAN’s parameter count of only 5.4 million and a storage requirement of 20 MB, alongside competitive image quality compared to Scenimefy. Furthermore, NijiGAN effectively reduces the occurrence of artifacts in the generated images. As shown in Figure \ref{checkeredartifacts}, NijiGAN eliminates the checkered artifact, producing images that are exceptionally smooth.

\begin{figure}[H]
  \centering
  \includegraphics[width=1\textwidth]{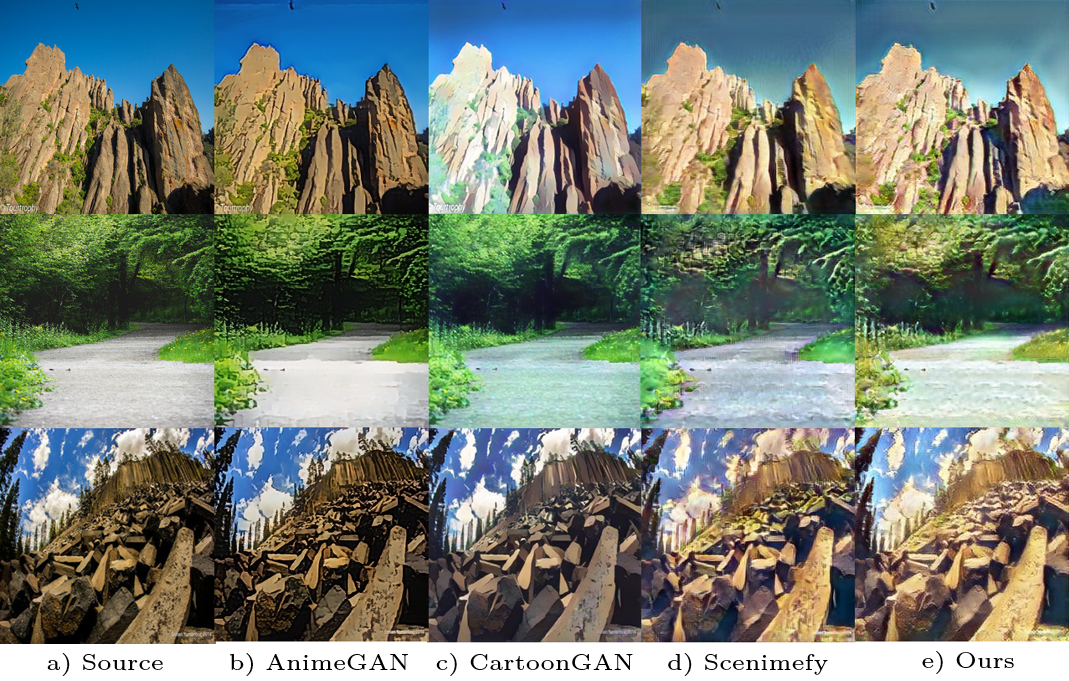}
  \caption{\textbf{Texture distribution of NijiGAN and several baseline models.} The anime style generated by each I2I models is illustrated.}\label{texturedistribution}
\end{figure}

In terms of texture distribution fidelity, the real image (leftmost) naturally has the best detail and sharpness as shown in Figure \ref{texturedistribution}. AnimeGAN simplifies textures significantly, leading to overly smoothed and cartoonish results with edge emphasis. CartoonGAN improves slightly but still flattens textures, losing finer details. Scenimefy performs better, retaining some natural texture distribution, though minor distortions and smoothing persist. NijiGAN stands out with the best balance, preserving fine details and textures while maintaining an anime-style aesthetic, making it the most faithful to the original image's texture distribution.

\section{Conclusion}\label{conclusion}
In this reseach, we propose an image-to-image translation model namely NijiGAN. NijiGAN performs the translation of real-world images into anime-style images by improving upon the reference model, Scenimefy, through the incorporation of NeuralODE in its generator. As a result, NijiGAN successfully achieves anime-style image translation with competitive quality compared to Scenimefy. By employing NeuralODE, the proposed model operates more efficiently in terms of parameter count and memory requirements, rendering NijiGAN remarkably lightweight and highly feasible for deployment on edge devices. However, NijiGAN exhibits a limitation in computational speed and TFLOPS performance due to its forward propagation. This drawback arises from the way NeuralODE models feature transformations within each block using continuous dynamics.





\bibliography{sn-bibliography}% common bib file


\begin{thebibliography}{24}
\ifx \bisbn   \undefined \def \bisbn  #1{ISBN #1}\fi
\ifx \binits  \undefined \def \binits#1{#1}\fi
\ifx \bauthor  \undefined \def \bauthor#1{#1}\fi
\ifx \batitle  \undefined \def \batitle#1{#1}\fi
\ifx \bjtitle  \undefined \def \bjtitle#1{#1}\fi
\ifx \bvolume  \undefined \def \bvolume#1{\textbf{#1}}\fi
\ifx \byear  \undefined \def \byear#1{#1}\fi
\ifx \bissue  \undefined \def \bissue#1{#1}\fi
\ifx \bfpage  \undefined \def \bfpage#1{#1}\fi
\ifx \blpage  \undefined \def \blpage #1{#1}\fi
\ifx \burl  \undefined \def \burl#1{\textsf{#1}}\fi
\ifx \doiurl  \undefined \def \doiurl#1{\url{https://doi.org/#1}}\fi
\ifx \betal  \undefined \def \betal{\textit{et al.}}\fi
\ifx \binstitute  \undefined \def \binstitute#1{#1}\fi
\ifx \binstitutionaled  \undefined \def \binstitutionaled#1{#1}\fi
\ifx \bctitle  \undefined \def \bctitle#1{#1}\fi
\ifx \beditor  \undefined \def \beditor#1{#1}\fi
\ifx \bpublisher  \undefined \def \bpublisher#1{#1}\fi
\ifx \bbtitle  \undefined \def \bbtitle#1{#1}\fi
\ifx \bedition  \undefined \def \bedition#1{#1}\fi
\ifx \bseriesno  \undefined \def \bseriesno#1{#1}\fi
\ifx \blocation  \undefined \def \blocation#1{#1}\fi
\ifx \bsertitle  \undefined \def \bsertitle#1{#1}\fi
\ifx \bsnm \undefined \def \bsnm#1{#1}\fi
\ifx \bsuffix \undefined \def \bsuffix#1{#1}\fi
\ifx \bparticle \undefined \def \bparticle#1{#1}\fi
\ifx \barticle \undefined \def \barticle#1{#1}\fi
\bibcommenthead
\ifx \bconfdate \undefined \def \bconfdate #1{#1}\fi
\ifx \botherref \undefined \def \botherref #1{#1}\fi
\ifx \url \undefined \def \url#1{\textsf{#1}}\fi
\ifx \bchapter \undefined \def \bchapter#1{#1}\fi
\ifx \bbook \undefined \def \bbook#1{#1}\fi
\ifx \bcomment \undefined \def \bcomment#1{#1}\fi
\ifx \oauthor \undefined \def \oauthor#1{#1}\fi
\ifx \citeauthoryear \undefined \def \citeauthoryear#1{#1}\fi
\ifx \endbibitem  \undefined \def \endbibitem {}\fi
\ifx \bconflocation  \undefined \def \bconflocation#1{#1}\fi
\ifx \arxivurl  \undefined \def \arxivurl#1{\textsf{#1}}\fi
\csname PreBibitemsHook\endcsname

%%% 1
\bibitem[\protect\citeauthoryear{Goodfellow et~al.}{2020}]{goodfellow2020generative}
\begin{barticle}
\bauthor{\bsnm{Goodfellow}, \binits{I.}},
\bauthor{\bsnm{Pouget-Abadie}, \binits{J.}},
\bauthor{\bsnm{Mirza}, \binits{M.}},
\bauthor{\bsnm{Xu}, \binits{B.}},
\bauthor{\bsnm{Warde-Farley}, \binits{D.}},
\bauthor{\bsnm{Ozair}, \binits{S.}},
\bauthor{\bsnm{Courville}, \binits{A.}},
\bauthor{\bsnm{Bengio}, \binits{Y.}}:
\batitle{Generative adversarial networks}.
\bjtitle{Communications of the ACM}
\bvolume{63}(\bissue{11}),
\bfpage{139}--\blpage{144}
(\byear{2020})
\end{barticle}
\endbibitem

%%% 2
\bibitem[\protect\citeauthoryear{Karras et~al.}{2019}]{karras2019style}
\begin{bchapter}
\bauthor{\bsnm{Karras}, \binits{T.}},
\bauthor{\bsnm{Laine}, \binits{S.}},
\bauthor{\bsnm{Aila}, \binits{T.}}:
\bctitle{A style-based generator architecture for generative adversarial networks}.
In: \bbtitle{Proceedings of the IEEE/CVF Conference on Computer Vision and Pattern Recognition},
pp. \bfpage{4401}--\blpage{4410}
(\byear{2019})
\end{bchapter}
\endbibitem

%%% 3
\bibitem[\protect\citeauthoryear{Karras et~al.}{2020}]{karras2020analyzing}
\begin{bchapter}
\bauthor{\bsnm{Karras}, \binits{T.}},
\bauthor{\bsnm{Laine}, \binits{S.}},
\bauthor{\bsnm{Aittala}, \binits{M.}},
\bauthor{\bsnm{Hellsten}, \binits{J.}},
\bauthor{\bsnm{Lehtinen}, \binits{J.}},
\bauthor{\bsnm{Aila}, \binits{T.}}:
\bctitle{Analyzing and improving the image quality of stylegan}.
In: \bbtitle{Proceedings of the IEEE/CVF Conference on Computer Vision and Pattern Recognition},
pp. \bfpage{8110}--\blpage{8119}
(\byear{2020})
\end{bchapter}
\endbibitem

%%% 4
\bibitem[\protect\citeauthoryear{Cohen and Giryes}{2023}]{cohen2023generative}
\begin{bchapter}
\bauthor{\bsnm{Cohen}, \binits{G.}},
\bauthor{\bsnm{Giryes}, \binits{R.}}:
\bctitle{Generative adversarial networks}.
In: \bbtitle{Machine Learning for Data Science Handbook: Data Mining and Knowledge Discovery Handbook},
pp. \bfpage{375}--\blpage{400}.
\bpublisher{Springer}, \blocation{???}
(\byear{2023})
\end{bchapter}
\endbibitem

%%% 5
\bibitem[\protect\citeauthoryear{Gui et~al.}{2021}]{gui2021review}
\begin{barticle}
\bauthor{\bsnm{Gui}, \binits{J.}},
\bauthor{\bsnm{Sun}, \binits{Z.}},
\bauthor{\bsnm{Wen}, \binits{Y.}},
\bauthor{\bsnm{Tao}, \binits{D.}},
\bauthor{\bsnm{Ye}, \binits{J.}}:
\batitle{A review on generative adversarial networks: Algorithms, theory, and applications}.
\bjtitle{IEEE transactions on knowledge and data engineering}
\bvolume{35}(\bissue{4}),
\bfpage{3313}--\blpage{3332}
(\byear{2021})
\end{barticle}
\endbibitem

%%% 6
\bibitem[\protect\citeauthoryear{Sharma et~al.}{2024}]{sharma2024generative}
\begin{botherref}
\oauthor{\bsnm{Sharma}, \binits{P.}},
\oauthor{\bsnm{Kumar}, \binits{M.}},
\oauthor{\bsnm{Sharma}, \binits{H.K.}},
\oauthor{\bsnm{Biju}, \binits{S.M.}}:
Generative adversarial networks (gans): Introduction, taxonomy, variants, limitations, and applications.
Multimedia Tools and Applications,
1--48
(2024)
\end{botherref}
\endbibitem

%%% 7
\bibitem[\protect\citeauthoryear{Bengesi et~al.}{2024}]{bengesi2024advancements}
\begin{botherref}
\oauthor{\bsnm{Bengesi}, \binits{S.}},
\oauthor{\bsnm{El-Sayed}, \binits{H.}},
\oauthor{\bsnm{Sarker}, \binits{M.K.}},
\oauthor{\bsnm{Houkpati}, \binits{Y.}},
\oauthor{\bsnm{Irungu}, \binits{J.}},
\oauthor{\bsnm{Oladunni}, \binits{T.}}:
Advancements in generative ai: A comprehensive review of gans, gpt, autoencoders, diffusion model, and transformers.
IEEE Access
(2024)
\end{botherref}
\endbibitem

%%% 8
\bibitem[\protect\citeauthoryear{Reed et~al.}{2016}]{reed2016generative}
\begin{bchapter}
\bauthor{\bsnm{Reed}, \binits{S.}},
\bauthor{\bsnm{Akata}, \binits{Z.}},
\bauthor{\bsnm{Yan}, \binits{X.}},
\bauthor{\bsnm{Logeswaran}, \binits{L.}},
\bauthor{\bsnm{Schiele}, \binits{B.}},
\bauthor{\bsnm{Lee}, \binits{H.}}:
\bctitle{Generative adversarial text to image synthesis}.
In: \bbtitle{International Conference on Machine Learning},
pp. \bfpage{1060}--\blpage{1069}
(\byear{2016}).
\bcomment{PMLR}
\end{bchapter}
\endbibitem

%%% 9
\bibitem[\protect\citeauthoryear{Zhang et~al.}{2022}]{zhang2022styleswin}
\begin{bchapter}
\bauthor{\bsnm{Zhang}, \binits{B.}},
\bauthor{\bsnm{Gu}, \binits{S.}},
\bauthor{\bsnm{Zhang}, \binits{B.}},
\bauthor{\bsnm{Bao}, \binits{J.}},
\bauthor{\bsnm{Chen}, \binits{D.}},
\bauthor{\bsnm{Wen}, \binits{F.}},
\bauthor{\bsnm{Wang}, \binits{Y.}},
\bauthor{\bsnm{Guo}, \binits{B.}}:
\bctitle{Styleswin: Transformer-based gan for high-resolution image generation}.
In: \bbtitle{Proceedings of the IEEE/CVF Conference on Computer Vision and Pattern Recognition},
pp. \bfpage{11304}--\blpage{11314}
(\byear{2022})
\end{bchapter}
\endbibitem

%%% 10
\bibitem[\protect\citeauthoryear{Kim and Park}{2020}]{kim2020remark}
\begin{barticle}
\bauthor{\bsnm{Kim}, \binits{M.}},
\bauthor{\bsnm{Park}, \binits{E.}}:
\batitle{A remark on convolution products for quiver hecke algebras}.
\bjtitle{International Journal of Mathematics}
\bvolume{31}(\bissue{11}),
\bfpage{2050092}
(\byear{2020})
\end{barticle}
\endbibitem

%%% 11
\bibitem[\protect\citeauthoryear{Zhu et~al.}{2017}]{zhu2017unpaired}
\begin{bchapter}
\bauthor{\bsnm{Zhu}, \binits{J.-Y.}},
\bauthor{\bsnm{Park}, \binits{T.}},
\bauthor{\bsnm{Isola}, \binits{P.}},
\bauthor{\bsnm{Efros}, \binits{A.A.}}:
\bctitle{Unpaired image-to-image translation using cycle-consistent adversarial networks}.
In: \bbtitle{Proceedings of the IEEE International Conference on Computer Vision},
pp. \bfpage{2223}--\blpage{2232}
(\byear{2017})
\end{bchapter}
\endbibitem

%%% 12
\bibitem[\protect\citeauthoryear{Liu et~al.}{2017}]{liu2017unsupervised}
\begin{botherref}
\oauthor{\bsnm{Liu}, \binits{M.-Y.}},
\oauthor{\bsnm{Breuel}, \binits{T.}},
\oauthor{\bsnm{Kautz}, \binits{J.}}:
Unsupervised image-to-image translation networks.
Advances in neural information processing systems
\textbf{30}
(2017)
\end{botherref}
\endbibitem

%%% 13
\bibitem[\protect\citeauthoryear{Huang et~al.}{2018}]{huang2018multimodal}
\begin{bchapter}
\bauthor{\bsnm{Huang}, \binits{X.}},
\bauthor{\bsnm{Liu}, \binits{M.-Y.}},
\bauthor{\bsnm{Belongie}, \binits{S.}},
\bauthor{\bsnm{Kautz}, \binits{J.}}:
\bctitle{Multimodal unsupervised image-to-image translation}.
In: \bbtitle{Proceedings of the European Conference on Computer Vision (ECCV)},
pp. \bfpage{172}--\blpage{189}
(\byear{2018})
\end{bchapter}
\endbibitem

%%% 14
\bibitem[\protect\citeauthoryear{Jiang et~al.}{2023}]{jiang2023scenimefy}
\begin{bchapter}
\bauthor{\bsnm{Jiang}, \binits{Y.}},
\bauthor{\bsnm{Jiang}, \binits{L.}},
\bauthor{\bsnm{Yang}, \binits{S.}},
\bauthor{\bsnm{Loy}, \binits{C.C.}}:
\bctitle{Scenimefy: learning to craft anime scene via semi-supervised image-to-image translation}.
In: \bbtitle{Proceedings of the IEEE/CVF International Conference on Computer Vision},
pp. \bfpage{7357}--\blpage{7367}
(\byear{2023})
\end{bchapter}
\endbibitem

%%% 15
\bibitem[\protect\citeauthoryear{Khrulkov et~al.}{2021}]{Khrulkov_2021_ICCV}
\begin{bchapter}
\bauthor{\bsnm{Khrulkov}, \binits{V.}},
\bauthor{\bsnm{Mirvakhabova}, \binits{L.}},
\bauthor{\bsnm{Oseledets}, \binits{I.}},
\bauthor{\bsnm{Babenko}, \binits{A.}}:
\bctitle{Latent transformations via neuralodes for gan-based image editing}.
In: \bbtitle{Proceedings of the IEEE/CVF International Conference on Computer Vision (ICCV)},
pp. \bfpage{14428}--\blpage{14437}
(\byear{2021})
\end{bchapter}
\endbibitem

%%% 16
\bibitem[\protect\citeauthoryear{Lee et~al.}{2024}]{LEE2024111170}
\begin{barticle}
\bauthor{\bsnm{Lee}, \binits{H.}},
\bauthor{\bsnm{Seol}, \binits{J.}},
\bauthor{\bsnm{Lee}, \binits{S.-g.}},
\bauthor{\bsnm{Park}, \binits{J.}},
\bauthor{\bsnm{Shim}, \binits{J.}}:
\batitle{Contrastive learning for unsupervised image-to-image translation}.
\bjtitle{Applied Soft Computing}
\bvolume{151},
\bfpage{111170}
(\byear{2024})
\doiurl{10.1016/j.asoc.2023.111170}
\end{barticle}
\endbibitem

%%% 17
\bibitem[\protect\citeauthoryear{Park et~al.}{2020}]{10.1007/978-3-030-58545-7_19}
\begin{bchapter}
\bauthor{\bsnm{Park}, \binits{T.}},
\bauthor{\bsnm{Efros}, \binits{A.A.}},
\bauthor{\bsnm{Zhang}, \binits{R.}},
\bauthor{\bsnm{Zhu}, \binits{J.-Y.}}:
\bctitle{Contrastive learning for unpaired image-to-image translation}.
In: \beditor{\bsnm{Vedaldi}, \binits{A.}},
\beditor{\bsnm{Bischof}, \binits{H.}},
\beditor{\bsnm{Brox}, \binits{T.}},
\beditor{\bsnm{Frahm}, \binits{J.-M.}} (eds.)
\bbtitle{Computer Vision -- ECCV 2020},
pp. \bfpage{319}--\blpage{345}.
\bpublisher{Springer},
\blocation{Cham}
(\byear{2020})
\end{bchapter}
\endbibitem

%%% 18
\bibitem[\protect\citeauthoryear{Han et~al.}{2021}]{Han_2021_CVPR}
\begin{bchapter}
\bauthor{\bsnm{Han}, \binits{J.}},
\bauthor{\bsnm{Shoeiby}, \binits{M.}},
\bauthor{\bsnm{Petersson}, \binits{L.}},
\bauthor{\bsnm{Armin}, \binits{M.A.}}:
\bctitle{Dual contrastive learning for unsupervised image-to-image translation}.
In: \bbtitle{Proceedings of the IEEE/CVF Conference on Computer Vision and Pattern Recognition (CVPR) Workshops},
pp. \bfpage{746}--\blpage{755}
(\byear{2021})
\end{bchapter}
\endbibitem

%%% 19
\bibitem[\protect\citeauthoryear{Yan et~al.}{2019}]{yan2019robustness}
\begin{botherref}
\oauthor{\bsnm{Yan}, \binits{H.}},
\oauthor{\bsnm{Du}, \binits{J.}},
\oauthor{\bsnm{Tan}, \binits{V.Y.}},
\oauthor{\bsnm{Feng}, \binits{J.}}:
On robustness of neural ordinary differential equations.
arXiv preprint arXiv:1910.05513
(2019)
\end{botherref}
\endbibitem

%%% 20
\bibitem[\protect\citeauthoryear{Chen et~al.}{2018}]{chen2018neural}
\begin{botherref}
\oauthor{\bsnm{Chen}, \binits{R.T.}},
\oauthor{\bsnm{Rubanova}, \binits{Y.}},
\oauthor{\bsnm{Bettencourt}, \binits{J.}},
\oauthor{\bsnm{Duvenaud}, \binits{D.K.}}:
Neural ordinary differential equations.
Advances in neural information processing systems
\textbf{31}
(2018)
\end{botherref}
\endbibitem

%%% 21
\bibitem[\protect\citeauthoryear{Kimura}{2009}]{kimura2009dormand}
\begin{barticle}
\bauthor{\bsnm{Kimura}, \binits{T.}}:
\batitle{On dormand-prince method}.
\bjtitle{Jpn. Malaysia Tech. Instit}
\bvolume{40},
\bfpage{1}--\blpage{9}
(\byear{2009})
\end{barticle}
\endbibitem

%%% 22
\bibitem[\protect\citeauthoryear{Jung et~al.}{2022}]{jung2022exploring}
\begin{bchapter}
\bauthor{\bsnm{Jung}, \binits{C.}},
\bauthor{\bsnm{Kwon}, \binits{G.}},
\bauthor{\bsnm{Ye}, \binits{J.C.}}:
\bctitle{Exploring patch-wise semantic relation for contrastive learning in image-to-image translation tasks}.
In: \bbtitle{Proceedings of the IEEE/CVF Conference on Computer Vision and Pattern Recognition},
pp. \bfpage{18260}--\blpage{18269}
(\byear{2022})
\end{bchapter}
\endbibitem

%%% 23
\bibitem[\protect\citeauthoryear{Andersson and Arvidsson}{2020}]{andersson2020generative}
\begin{botherref}
\oauthor{\bsnm{Andersson}, \binits{F.}},
\oauthor{\bsnm{Arvidsson}, \binits{S.}}:
Generative adversarial networks for photo to hayao miyazaki style cartoons.
arXiv preprint arXiv:2005.07702
(2020)
\end{botherref}
\endbibitem

%%% 24
\bibitem[\protect\citeauthoryear{Chen et~al.}{2020}]{chen2020animegan}
\begin{bchapter}
\bauthor{\bsnm{Chen}, \binits{J.}},
\bauthor{\bsnm{Liu}, \binits{G.}},
\bauthor{\bsnm{Chen}, \binits{X.}}:
\bctitle{Animegan: A novel lightweight gan for photo animation}.
In: \bbtitle{Artificial Intelligence Algorithms and Applications: 11th International Symposium, ISICA 2019, Guangzhou, China, November 16--17, 2019, Revised Selected Papers 11},
pp. \bfpage{242}--\blpage{256}
(\byear{2020}).
\bcomment{Springer}
\end{bchapter}
\endbibitem

\end{thebibliography}
% \input NijiGAN.bbl

\end{document}